# XTAG System - A Wide Coverage Grammar for English

Christy Doran, Dania Egedi, Beth Ann Hockey, B. Srinivas and Martin Zaidel*

Institute for Research in Cognitive Science
University of Pennsylvania
Philadelphia, PA 19104-6228, USA
{cdoran, egedi, beth, srini, zaidel}@linc.cis.upenn.edu

## Abstract

This paper presents the XTAG system, a grammar development tool based on the Tree Adjoining Grammar (TAG) formalism that includes a wide-coverage syntactic grammar for English. The various components of the system are discussed and preliminary evaluation results from the parsing of various corpora are given. Results from the comparison of XTAG against the IBM statistical parser and the Alvey Natural Language Tool parser are also given.

## 1 INTRODUCTION

XTAG is a large on-going project to develop a wide-coverage grammar for English, based on the Lexicalized Tree Adjoining Grammar (LTAG) formalism. LTAG is a lexicalized mildly-context sensitive tree rewriting system [Joshi *et al.*, 1975; Schabes, 1990] that is closely related to Dependency Grammars and Categorial Grammars. Elementary trees in LTAG provide a larger domain of locality over which syntactic and semantic (predicate-argument) constraints are specified. XTAG also serves as an LTAG grammar development system consisting of a predictive left-to-right parser, an X-window interface, a morphological analyzer, and a part-of-speech tagger (also referred to as simply 'tagger').

## 2 SYSTEM DESCRIPTION

Figure 1 shows the overall flow of the system when parsing a sentence. The input sentence is submitted to the **Morphological Analyzer** and the **Tagger**. The morphological analyzer retrieves the morphological information for each individual word from the morphological database. This output is filtered in the **P.O.S Blender** using the output of the trigram tagger to reduce the part-of-speech ambiguity of the words. The sentence, now annotated with part-of-speech tags and morphological information for each word, is input to the **Parser**, which consults the syntactic database and tree database to retrieve the appropriate tree structures for each lexical item. A variety of heuristics are used to reduce the number of trees selected. The parser then composes the structures to obtain the parse(s) of the sentence.

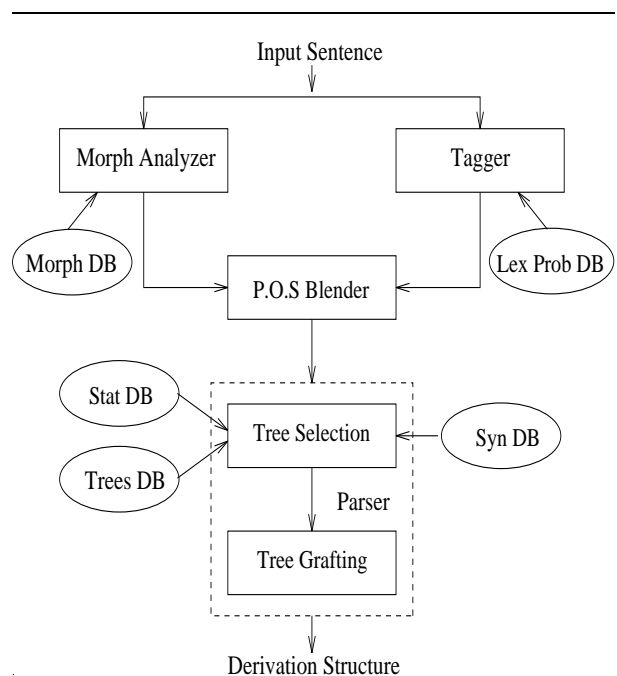

Figure 1: **Overview of XTAG system**

### 2.1 Morphological Analyzer

The morphology database [Karp *et al.*, 1992] was originally extracted from 1979 edition of the Collins English Dictionary and Oxford Advanced Learner's

---

*currently at BBN, Cambridge, MA, USA

Dictionary of Current English, and then cleaned up and augmented by hand. It consists of approximately 317,000 inflected items, along with their root forms and inflectional information (such as case, number, tense). Thirteen parts of speech are differentiated: Noun, Proper Noun, Pronoun, Verb, Verb Particle, Adverb, Adjective, Preposition, Complementizer, Determiner, Conjunction, Interjection, and Noun/Verb Contraction. Nouns and Verbs are the largest categories, with approximately 213,000 and 46,500 inflected forms, respectively. The access time for a given inflected entry is 0.6 msec.

## 2.2 Part-of-Speech Tagger

A trigram part-of-speech tagger [Church, 1988], trained on the Wall Street Journal Corpus, is incorporated in XTAG. The trigram tagger has been extended to output the N-best parts-of-speech sequences [Soong and Huang, 1990]. XTAG uses this information to reduce the number of specious parses by filtering the possible parts-of-speech provided by the morphological analyzer for each word. The tagger decreases the time to parse a sentence by an average of 93%.

## 2.3 Parser

The system uses an Earley-style parser which has been extended to handle feature structures associated with trees [Schabes, 1990]. The parser uses a general two-pass parsing strategy for 'lexicalized' grammars [Schabes, 1988]. In the tree-selection pass, the parser uses the syntactic database entry for each lexical item in the sentence to select a set of elementary structures from the tree database. The tree-grafting pass composes the selected trees using substitution and adjunction operations to obtain the parse of the sentence. The output of the parser for the sentence *I had a map yesterday* is illustrated in Figure 2. The **parse tree**[1] represents the surface constituent structure, while the **derivation tree** represents the derivation history of the parse. The nodes of the derivation tree are the tree names anchored by the lexical items. The composition operation is indicated by the nature of the arcs; a dashed line is used for substitution and a bold line for adjunction. The number beside each tree name is the address of the node at which the operation took place. The derivation tree can also be interpreted as a dependency graph with unlabeled arcs between words of the sentence.

Heuristics that take advantage of LTAGs have been implemented to improve the performance of the parser. For instance, the span of the tree and the position of the anchor in the tree are used to weed out unsuitable trees in the first pass of the parser. Statistical information about the usage frequency of the trees has been acquired by parsing corpora. This information has been compiled into a statistical database (the **Lex Prob DB**) that is used by the parser. These methods speed the runtime by approximately 87%.

---

[1]Each node of the parse tree has a feature structure, not shown here, associated with it.

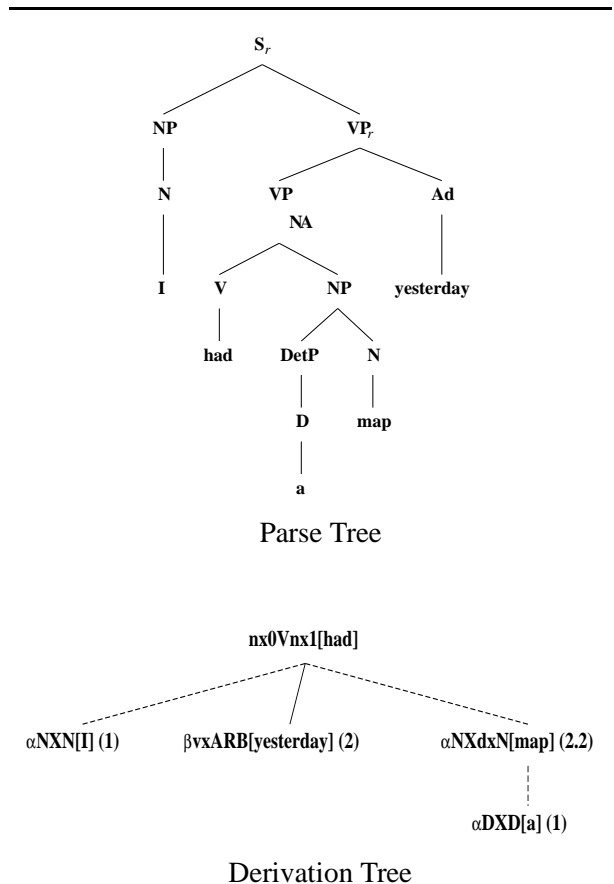

Figure 2: Output structures from the Parser

### 2.3.1 Heuristics for Ranking the Parses

The parser generates the parses in a rank order. This ranking is determined using a combination of heuristics, which are expressed as structural preferences for derivation, e.g. attachment sites of adjuncts, right- vs. left- branching structures, topicalized sentences, etc. Similar heuristics have been used for other parsers. See recent work by [Hobbs and Bear, 1994], [McCord, 1993],and [Nagao, 1994].

A partial list of the heuristics used in XTAG follows:

1. Prefer argument positions to adjunct positions (here, this amounts to preferring fewer adjunction operations).

2. For PPs other than *of*, attach to nearest site that is not a proper noun.

3. Prefer right-branching structure for sequences of adjectives, adverbs, and PPs.

4. Prefer left-branching structure for sequences of nouns.

5. Prefer high attachment (wide-scope) for a modifier and a sequence of modifiees of the same type (i.e. a PP following or preceding a coordinate NP, an adjective or determiner preceding a coordinate NP or sequence of Ns, an N preceding coordinate Ns).

These rankings are used to control the number of sentences passed on to further levels of processing. In applications emphasizing speed, only the highest ranked parse will be considered. In applications emphasizing accuracy, the top N parses can be considered.

### 2.3.2 Syntactic Database

The syntactic database associates lexical items with the appropriate trees and tree families based on selectional information. The syntactic database entries were originally extracted from the Oxford Advanced Learner's Dictionary and Oxford Dictionary for Contemporary Idiomatic English, and then modified and augmented by hand. There are more than 37,000 syntactic database entries. Selected entries from this database are shown in Table 1. Each syntactic entry consists of an INDEX field, the uninflected form under which the entry is compiled in the database, an ENTRY field, which contains all of the lexical items that will anchor the associated tree(s), a POS field, which gives the part-of-speech for the lexical item(s) in the ENTRY field, and then either (but not both) a TREES or FAM field. The TREES field indicates a list of individual trees to be associated with the entry, while the FAM field indicates a list of tree families. A tree family, which corresponds to a subcategorization frame (see section 2.3.3), may contain a number of trees. A syntactic entry may also contain a list of feature templates (FS) which expand out to feature equations to be placed in the specified tree(s). Any number of EX fields may be provided for example sentences. Note that lexical items may have more than one entry and may select the same tree more than once, using different features to capture lexical idiosyncrasies (e.g. *have*).

| | |
|---|---|
| INDEX: | have/26 |
| ENTRY: | have |
| POS: | V |
| TREES: | $\beta$Vvx |
| FS: | #VPr_ind, #VPr_past, #VPr_perfect+ #VP_ppart, #VP_pass- |
| EX: | he had died; we had died |
| | |
| INDEX: | have/50 |
| ENTRY: | have |
| POS: | V |
| TREES: | $\beta$Vvx |
| FS: | #VP_inf |
| EX: | John has to go to the store. |
| | |
| INDEX: | have/69 |
| ENTRY: | NP0 have NP1 |
| POS: | NP0 V NP1 |
| FAM: | Tnx0Vnx1 |
| FS: | #TRANS+ |
| EX: | John has a problem. |
| | |
| INDEX: | map/1 |
| ENTRY: | NP0 map out NP1 |
| POS: | NP0 V PL NP1 |
| FAM: | Tnx0Vplnx1 |
| | |
| INDEX: | map/3 |
| ENTRY: | map |
| POS: | N |
| TREES: | $\alpha$N, $\alpha$NXdxN, $\beta$Nn |
| FS: | #N_wh-, #N_refl- |
| | |
| INDEX: | map/4 |
| ENTRY: | map |
| POS: | N |
| TREES: | $\alpha$NXN |
| FS: | #N_wh-, #N_refl-, #N_plur |

Table 1: Selected Syntactic Database Entries

### 2.3.3 Tree Database

Trees in the English LTAG framework fall into two conceptual classes. The smaller class consists of individual trees such as trees (a), (d), and (e) in Figure 3. These trees are generally anchored by non-verbal lexical items. The larger class consists of trees that are grouped into tree families. These tree families represent subcategorization frames; the trees in a tree family would be related to each other transformationally in a movement-based approach. Trees 3(b) and 3(c) are members of two distinct tree families. As illustrated by trees 3(d) and 3(e), each node of a tree is annotated with a set of features whose values may be specified within the tree or may be derived from the syntactic database. There are 385 trees that compose 40 tree families, along with 62 individually selected trees in the tree database.

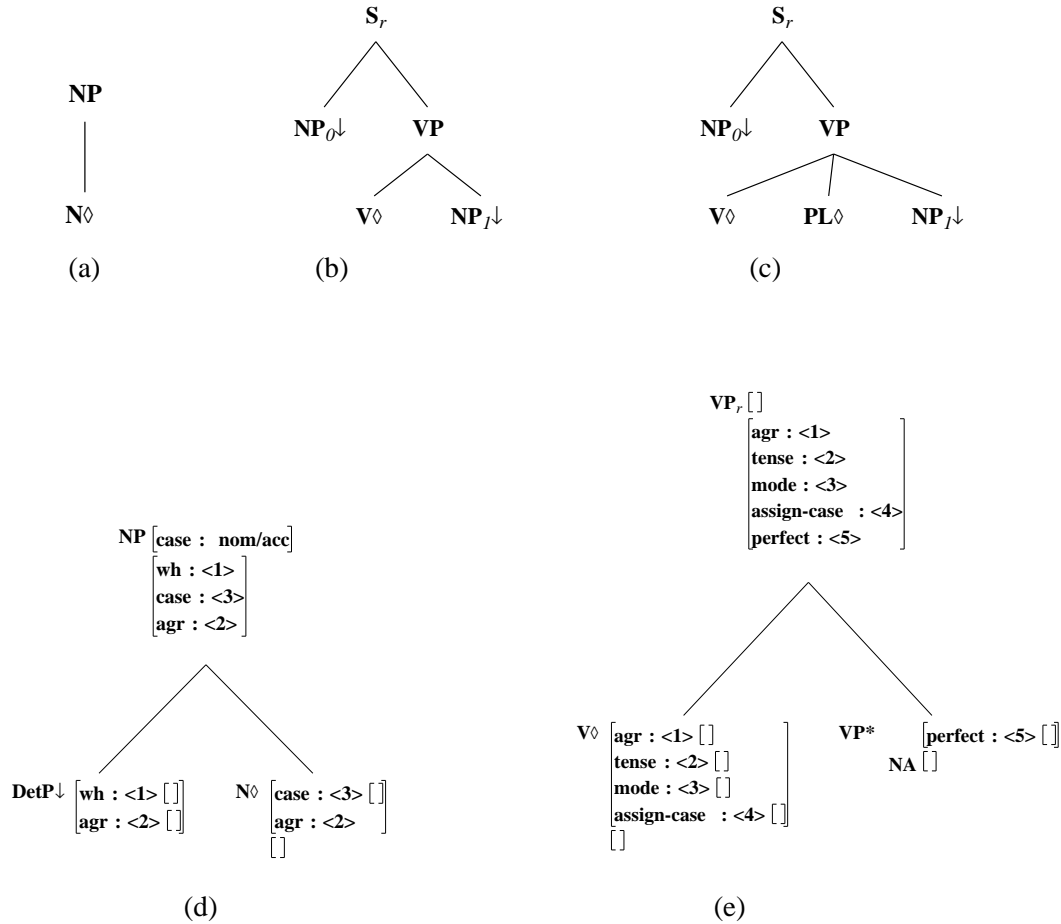

Figure 3: Sample trees from the Tree Database

### 2.3.4 Statistics Database

The statistics database contains tree unigram frequencies which have been collected by parsing the Wall Street Journal, the IBM manual, and the ATIS corpus using the XTAG English grammar. The parser, augmented with the statistics database [Joshi and Srinivas, 1994], assigns each word of the input sentence the top three most frequently used trees given the part-of-speech of the word. On failure the parser retries using all the trees suggested by the syntactic database for each word. The augmented parser has been observed to have a success rate of 50% without retries.

### 2.4 X-Interface

XTAG provides a graphical interface for manipulating TAGs. The interface offers the following:

- Menu-based facility for creating and modifying tree files and loading grammar files.
- User controlled parser parameters, including the parsing of categories (S, embedded S, NP, DetP), and the use of the tagger (on/off/retry on failure).
- Storage/retrieval facilities for elementary and parsed trees as text files.
- The production of postscript files corresponding to elementary and parsed trees.
- Graphical displays of tree and feature data structures, including a scroll 'web' for large tree structures.
- Mouse-based tree editor for creating and modifying trees and feature structures.
- Hand combination of trees by adjunction or substitution for use in diagnosing grammar problems.

Figure 4 shows the X window interface after a number of sentences have been parsed.

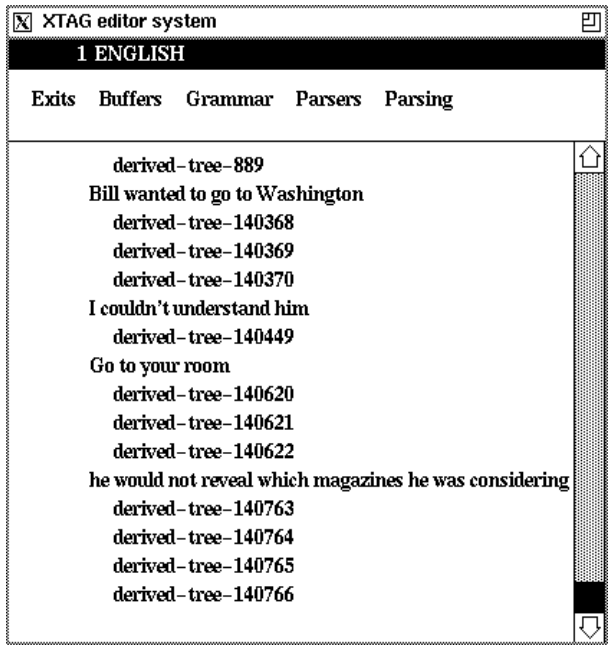

Figure 4: **XTAG Interface**

## 3 ENGLISH GRAMMAR

The morphology, syntactic, and tree databases together comprise the English grammar. Lexical items not in the databases are handled by default mechanisms. The range of syntactic phenomena that can be handled is large and includes auxiliaries (including inversion), copula, raising and small clause constructions, topicalization, relative clauses, infinitives, gerunds, passives, adjuncts, it-clefts, wh-clefts, PRO constructions, noun-noun modifications, extraposition, determiner phrases, genitives, negation, noun-verb contractions and imperatives. Analyses for sentential adjuncts and time NP adverbials are currently being implemented. The combination of large scale lexicons and wide phenomena coverage result in a robust system.

## 4 CORPUS PARSING AND EVALUATION

XTAG has recently been used to parse the Wall Street Journal[2], IBM manual, and ATIS corpora as a means of evaluating the coverage and correctness of XTAG parses. For this evaluation, a sentence is considered to have parsed correctly if XTAG produces parse trees. Verifying the presence of the correct parse among the parses generated is done manually at present. Table 2 shows the preliminary results. We will present more complete and rigorous results by the time of the conference and compare them with other parsers in the same class as XTAG. Although XTAG is being extended to handle sentence fragments, they are not included at present, and are thereby not reflected in the data in Table 2. Statistical information from the parsed corpora described in Section 2.3.4 is presently used only for speeding the parser but not to tune the grammar to parse any specific corpus. Note then, that the data below does not involve any corpus training.

| Corpus | # of Sents | % Parsed | Av. # of parses/sent |
|---|---|---|---|
| WSJ | 6364 | 39.09% | 7.53 |
| IBM Manual | 1611 | 75.42% | 6.14 |
| ATIS | 649 | 74.42% | 6.0 |

Table 2: Performance of XTAG on various corpora

### 4.1 Comparison with IBM Parser

A more detailed experiment to measure the crossing bracket accuracy of the XTAG-parsed IBM-manual sentences has been performed. Of the 1600 IBM sentences that have been parsed (those available from the Penn Treebank [Marcus *et al.*, 1993]), only 67 overlapped with the IBM-manual treebank that was bracketed by University of Lancaster.[3] The XTAG-parses for these 67 sentences were compared[4] with the Lancaster IBM-manual treebank.

Table 3 shows the results obtained in this experiment. It also shows the crossing bracket accuracy of the latest IBM statistical parser [Jelinek *et al.*, 1994] on the same genre of sentences. Recall is a measure of the number of bracketed constituents the system got right divided by the number of constituents in the corresponding Treebank sentences. Precision is the number of bracketed constituents the system got right divided by the number of bracketed constituents in the system's parse.

Based on the present data, we believe our results will be consistent for the complete XTAG-parsed IBM corpus; we plan to evaluate the XTAG parses against the Penn Treebank. In addition, we are parsing the Lancaster sentences, and adding those to the XTAG IBM corpus.

While the crossing-brackets measure is useful for comparing the output of different parsers, we believe that it is a somewhat inadequate method for evaluating a parser like XTAG for two main reasons. First, the parse generated by the XTAG system is much richer in its representation of the internal structure of certain phrases than those present in manually created treebanks. Even though the Lancaster treebank is more detailed in terms of bracketing than the Penn Treebank, it is not complete in its bracketing of the internal structure of noun phrases. As a result of comparing the XTAG parse with a skeletal representation, the precision score is misleadingly low for the XTAG system.

---

[2]Sentences of length $<=$ 15 words

[3]The treebank was obtained through Salim Roukos (roukos@watson.ibm.com) at IBM.

[4]We used the parseval program written by Phil Harison (phil@atc.boeing.com).

| System | # of sents | Crossing Brackets | Recall | Precision |
|---|---|---|---|---|
| XTAG | 67 | 80% | 84.32% | 59.28% |
| IBM Stat. parser | 1000 | 86.2% | Not Available | Not Available |

Table 3: Performance of XTAG on IBM-manual sentences

| System | # parsed | % parsed | Max dervs | Av dervs |
|---|---|---|---|---|
| ANLT Parser | 127 | 88.81% | 32 | 4.57 |
| XTAG with POS tagger | 93 | 65.03% | 28 | 3.45 |
| XTAG without POS tagger | 124 | 86.71% | 28 | 4.14 |

Table 4: Comparison of XTAG and ANLT Parser

A second reason that the crossing bracket measure is inadequate for evaluating XTAG is that the primary structure in XTAG is the derivation tree from which the bracketed tree is derived. Two identical bracketings for a sentence can have completely different derivation trees. A more direct measure of the performance of the XTAG parser would evaluate the derivation structure, which captures the dependencies between words.

### 4.2 Comparison with Alvey

We also compared the XTAG parser to the Alvey Natural Language Tools (ANLT) Parser, and found that the two performed comparably. We parsed the set of LDOCE Noun Phrases presented in Appendix B of the technical report [Carroll, 1993], using the XTAG parser. The technical report presents the ranking of the correct parse and also gives the total number of derivations for each noun phrase. In this experiment, we have compared the total number of derivations obtained from XTAG with that obtained from the ANLT parser.

Table 4 summarizes the results of this experiment. A total of 143 noun phrases were parsed. The NPs which did not have a correct parse in the top three derivations for the ANLT parser were considered as failures for ANLT. The maximum and average number of derivations columns show the highest and the average number of derivations produced for the NPs that have a correct derivation in the top three derivations. For the XTAG system, performance results with and without the POS tagger are shown.[5]

It would be interesting to see if the two systems performed similarly on a wider range of data. In [Carroll, 1993], only the LDOCE NPs are annotated with the number of derivations; we are interested in getting more data annotated with this information, in order to make further comparisons.

## 5 COMPUTER PLATFORM

XTAG was developed on the Sun SPARC station series, and has been tested on the Sun 4 and HP BOBCATs series 9000. It is available through anonymous ftp, and requires 20MB of space. Please send mail to xtag-request@linc.cis.upenn.edu for ftp instructions or more information. XTAG requires the following software to run:

- A machine running UNIX and X11R4. Previous releases of X will not work. X11R4 is free software available from MIT.

- A Common Lisp compiler which supports the latest definition of Common Lisp (Steele's Common Lisp, second edition). XTAG has been tested with Lucid Common Lisp 4.0 and Allegro 4.0.1.

- CLX version 4 or higher. CLX is the lisp equivalent to the Xlib package written in C.

- Mark Kantrowitz's Lisp Utilities from CMU: logical-pathnames and defsystem.

The latest version of CLX (R5.0) and the CMU Lisp Utilities are provided in our ftp directory for your convenience. However, we ask that you refer to the appropriate source for updates.

The morphology database component[Karp *et al.*, 1992], no longer under licensing restrictions, is available as a separate system from the XTAG system. Ftp instructions and more information can be obtained by mailing requests to lex-request@linc.cis.upenn.edu.

---

[5]Because the NPs are, on average, shorter than the sentences on which it was trained, the performance of the POS tagger is significantly degraded. For this reason, the table contains information about the performance of XTAG both with and without the tagger.